# Site-Specific Color Features of Green Coffee Beans

Shu-Min Tan[1,¶], Shih-Hsun Hung[1,¶], Je-Chiang Tsai[1,2,¶]

## Abstract

Coffee is one of the most valuable primary commodities. Despite this, the common selection technique of green coffee beans relies on personnel visual inspection, which is labor-intensive and subjective. Therefore, an efficient way to evaluate the bean's quality is needed. In this paper, we demonstrate a site-independent approach to find site-specific color features of the seed coat in qualified green coffee beans. We then propose two evaluation schemes for green coffee beans based on this site-specific color feature of qualified beans. Due to the site-specific properties of these color features, machine learning classifiers indicate that compared with the existing evaluation schemes of beans, our evaluation schemes have the advantages of being simple, having less computational costs, and having universal applicability. Finally, this site-specific color feature can distinguish qualified beans from different growing sites. Moreover, this function can prevent cheating in the coffee business and is unique to our evaluation scheme of beans.

## 1. Introduction

Coffee is one of the most important agricultural commodities in commodity markets. The market price of coffee strongly depends on the quality of green coffee beans. Despite the increased demand for high-quality coffee beans, the conventional selection technique mainly relies on personnel visual inspection, which is subjective and time-consuming. Thus, an efficient and reliable system for evaluating the quality of green coffee beans is needed. This system should be affordable to small-scale farmers or local coffee bean producers in developing countries, both technologically and in price.

[1] Department of Mathematics, National Tsing Hua University, Hsinchu 300, Taiwan. [2] Department of Mathematics, National Center for Theoretical Science, National Taiwan University, Taipei 106, Taiwan.
[¶] These authors contributed equally to this work. Corresponding author: Je-Chiang Tsai (Email address: tsaijc.math@gmail.com).

Another option for evaluating the quality of coffee beans is mechanical; a better mechanic evaluating option is the computer vision system, which can identify the defects of coffee beans without damaging them. The deficiency of green coffee beans can be classified into two types: Morphological defects, such as the shape and size of beans, and color defects, which are undesirable colors, such as brown to black, light brown to dark brown, and amber to yellow [1]. Morphological defects are associated with broken, insect-damaged, shell, and long berry beans, while color defects are with fade, black, fermented, moldy, insect-damaged, and sour beans [1-3] (see Fig. 1 and **Supplementary material** for more details). Fade, black, fermented, moldy, and sour defects are believed to be associated with chemical changes [4, 5].

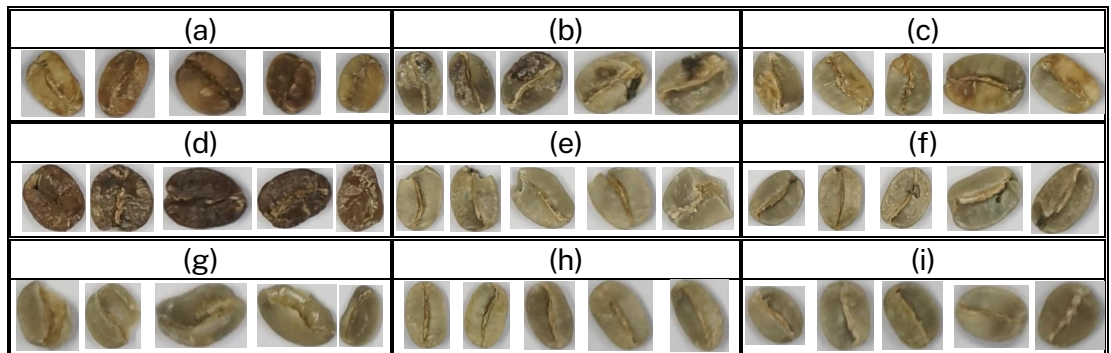

**Fig. 1 | Examples of defective green coffee beans[1-3].** (a) Fermented defect type. (b) Moldy defect type. (c) Fade defect type. (d) Black defect type. (e) Broken defect type. (f) Insect-damaged defect type. (g) Shell defect type. (h) Long berry defect type. (i) Normal beans.

Both morphological and color defects can be identified using image analysis. On the other hand, the imaging technique for identifying beans with morphological deficiency is well-developed due to the existence of the digital relations characterizing the morphological features such as the total surface area of beans (for the distinction of small beans), the roundness of beans, the ratio between the damaged area and the total surface area of beans, as well as the eccentricity value (for the distinction of very long berry and broken beans) [3, 6]. However, due to the presence of multiple color models, such as the RGB model, HSV model (Hue, Saturation, and Value), and CIE model, there is no commonly accepted color characteristic for distinguishing coffee beans. So the classification of coffee beans based on color defects is still being further explored. The framework of computer vision algorithms based on color defects involves extracting the color features of the image of coffee beans and then choosing a suitable machine learning classifier based on the extracted feature, such as a Support Vector

Machine or K-nearest neighbor method, to distinguish beans (see Garcia et al. [3]). The color characteristic can be all of the RGB (Red Green Blue) values of each pixel in the image of a coffee bean which would result in a large number of variables in the machine learning classifier, and thus the need for high-performance computing instruments or the cost of time-consuming [7]. The color characteristic of beans can be the RGB values of the center point of the image of a bean, and normal beans are identified by checking whether their RGB characteristic values are within the admissible range determined by the images of the training set of beans (see Arboleda, Fajard, Medina [8]). In addition to the RGB values of beans, there are two related color attributes for the classification of beans: CIE (Commission Internationale d'Eclairage) L*a*b* values and spectral values. de Oliveira et al. [9] (also see Krishna and Chakravarthy [10]) developed an approach that converts device-dependent RGB color matrixes used by a digital camera to the device-independent CIE L*a*b* color values using Artificial Neural Networks and then classifies green coffee beans based on their measured CIE L*a*b color units and the use of Bayesian classifier. This approach needs an auxiliary transformation model and, thus, the additional computational cost. Chen et al. [11] identified the spectral signatures of defective beans, which, together with deep learning data processing 2D- 3D merged CNN (Convolution neural network), can be employed to design a real-time coffee bean defect inspection system. Although this system has high accuracy, the construction of the system is complex and costly. Hendrawan et al. [12] analyzed 286 image features to obtain three image features that best characterize three types of local Indonesian green coffee beans. However, these three features may not characterize beans from other growing sites. We remark that there are other subjects for the coffee bean classification, for example, the identification of the coffee roasting degree (see Leme et al. [13]). Also, see Ligar [14] for a review of the computer vision method for identifying qualified beans.

    In the existing literature, such as the aforementioned ones, either the color feature is not a site-specific one to beans or complex image pre-processing is needed. Also, the efficiency of machine learning classifiers strongly depends on the chosen feature of beans, not the classifiers themselves. Although deep learning data processing (e.g., convolutional neural network) can extract the features of coffee beans by itself, it is time-consuming [14]. Moreover, the obtained feature may not be site-specific to beans, meaning it may not be present in beans from other groups of the same growing sites. Further, if the obtained feature of beans is not site-specific and the number of beans (in the training set) is not sufficiently large, the obtained feature can only apply to a

restricted class of beans [1]. In conclusion, the more site-specific the feature of the bean is, the more accurate the classification will be, the smaller the training set will be needed, and thus the lower the computational cost of the identifying scheme. In this paper, we demonstrate a set of site-specific color features of qualified green coffee beans. With the use of this set of site-specific color features, we can employ the Support Vector Machine classifier [15] (with the small ratio of the size of the training set to that of the whole set being 0.05) to classify beans into normal and defective beans, with an accuracy between 82.28% and 97.10% for beans from four different varieties or growing sites (see Table 3). Therefore, even with a very small amount of information, our identifying scheme can be accurate, indicating that our proposed color features of beans are truly site-specific. Further, due to the site-specific and the simplicity of our proposed features, our evaluation scheme of beans has the advantage of the small size of training sets, universal applicability to beans from different varieties or production sites and low-cost computation. Therefore, our bean evaluation scheme can be embedded in a low-cost and easy-operate hardware system, which would thus be easily introduced to small-scale farmers and bean producers to increase their production significantly.

## 2. Materials and Methods

### 2.1. Green Coffee Bean Samples

There are two major types of coffee beans: Arabica and Robusta. In this study, according to their grown locations, four classes of sample beans are prepared, as stated in the following.

1. Arabica beans were grown in Taiwu Mountain (high altitude of 1000 meters), Pingtung County, Taiwan. The total number of sample beans from this area is 600, including 300 normal beans and 300 defective beans.
2. Arabica beans were grown in Alishan Mountain (high altitude of 1100 meters), Chiayi County, Taiwan. The total number of sample beans from this area is 1920, including 960 normal beans and 960 defective beans.
3. Arabica beans (Typica, Yellow Bourbon, Venecia, Purpurascens) were grown in Kantou Mountain (high altitude of 700-750 meters), Tainan City, Taiwan. The total number of sample beans from this area is 600, including 300 normal beans and 300 defective beans.

4. Robusta beans were grown in Karnataka, (high altitude of 900-1000 meters), India. The total number of sample beans from this area is 436, including 218 normal beans and 218 defective beans.

## 2.2. Image Acquisition

The digital image is captured by a SONY NEX-F3 camera configured at the top of a mini-studio box which is equipped with LED lights and is made of black hard paper to eliminate the reflective light. The bottom of the box is placed with white paper and beans on it, which facilitates the separation of beans from the background. The overall shooting environment is shown in Fig. 2. The LED lights with illumination controllers are set on the top of the mini-studio box. The feasible region of the background for capturing coffee beans is about $20 \times 20 \times 30$ cm$^3$, allowing us to collect 64 coffee beans in one snapshot. To avoid specular reflection, the uniformity of the illumination is tested using a colorimeter before the acquisition step. Also, the differences in illumination before and after placing the coffee beans are tested, and the result shows that placing good beans, bad beans, or both on the background would not change the uniformity of the illumination significantly. These images were stored in JPEG (Joint Photographic Expert Group) format with a size of about $4912 \times 2760$ pixels.

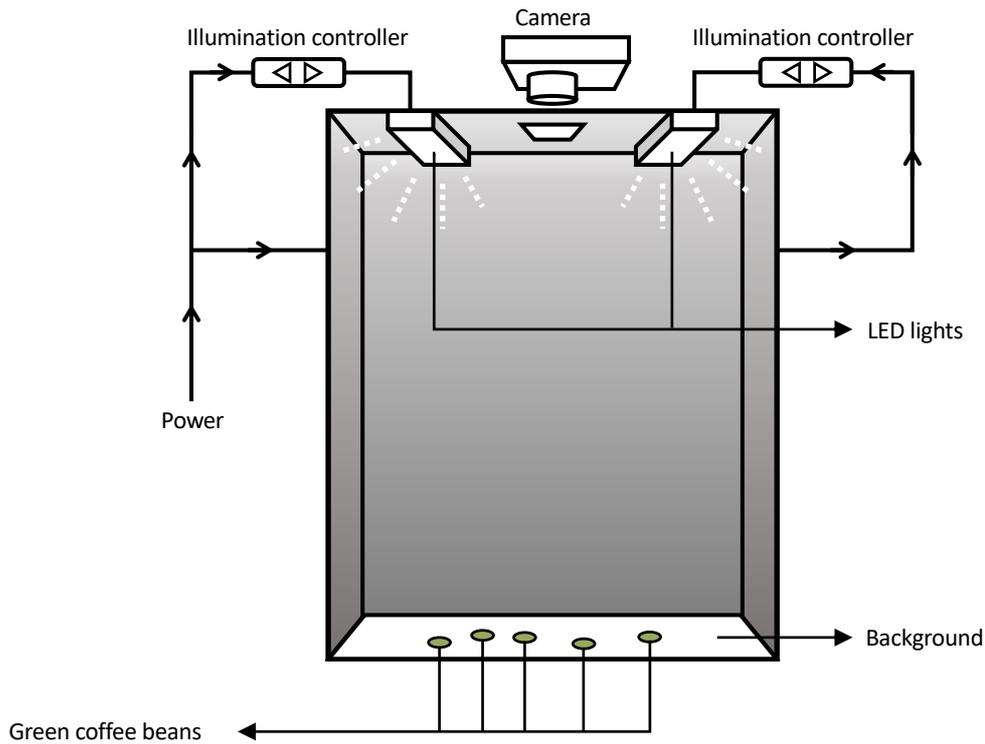

**Figure 2. The shooting environment.** The mini studio box is about 20 × 20 × 30 cm$^3$. The box is equipped with LED lights on top and external illumination controllers.

## 2.3. Image Processing

### 2.3.1. Basic materials for grayscale and RGB images [16]

An image is made up of a large amount of tiny and square-like elements, which are called pixels. A grayscale image has only one color channel. The pixel value of a pixel in a grayscale image represents the amount of gray intensity to be displayed for this specific pixel of the image. Each pixel value in a grayscale image is a single integer ranging from 0 to 255. The pixel value 0 represents black, while 255 represents white. On the other hand, an RGB image consists of three primary color channels: Red channel, Green channel, and Blue channel. In an RGB image, the pixel value of a pixel is represented by a triple (r, g, b) where the values r, g, and b stand for the intensity of red, green, and blue colors, respectively, needed to render this specific pixel of the RGB image. As in the grayscale image, each component of the pixel value (r, g, b) is an integer ranging from 0 to 255. The RGB color space is the set of all such possible triples (r, g, b). For example, the pixel value (255, 255, 0) represents a yellow color pixel — there is 100% red and green color in the pixel while there is no blue color involved in the pixel. A suitable combination of red, green, and blue intensities can give rise to any desired color. An RGB image can be converted into a primary color grayscale image by retaining any one of its three channels.

### 2.3.2. Image Segmentation

Image segmentation is to extract the images of beans from the white background of the acquired image. The segmentation is achieved by first converting the acquired color image to a grayscale image by retaining a specific primary color channel in RGB color space, then analyzing the histogram of pixel values of the grayscale image to choose a threshold value for distinguishing the background and the beans. Finally, based on whether each of the three primary color grayscale values of the image's pixel is larger than the chosen threshold, which is 163 in this paper, the pixel is classified into background and beans otherwise. For example, a pixel with RGB values (120, 170, 85) is classified into one part of a bean, while a pixel with RGB values (180, 170, 205) is classified into one part of the background. After segmentation, the image of a single bean is with a size of about 35 × 45 pixels.

### 2.4. Methodology of our identifying scheme: Supervised learning algorithm

Now, we briefly state the methodology of our identifying scheme of beans. We will propose two identifying schemes of beans: the first uses two statistical characteristics of beans and the other six statistical characteristics. The second scheme has a higher accuracy rate than the first one. The goal of the identifying scheme is to take a vector $x \in \mathbb{R}^n$ associated with a bean ( $n = 2$ or $6$ ), which consists of color features/characteristics of a bean, and give the value of a scalar $z \in \mathbb{R}$ as its output, which stands for a qualified bean if $z$ is smaller than the threshold 0 (when $n = 2$), and a defective bean otherwise. To do this, we divide the beans into two groups: the training set and the test set. We take $z$ to be a linear function of the input $x \in \mathbb{R}^n$, as given by

$$z = w^T x - b \qquad (1)$$

where $w \in \mathbb{R}^n$ is a vector of weight parameter, $w^T$ is the transpose of the vector $w$, and $b$ is a scalar weight parameter, with the property that almost every $z$ associated with a qualified bean and those associated with a defective bean lies in the two different regions separated by a hyperplane $E$ in $\mathbb{R}^n$ determined by equation (1). The optimized pair $(w, b)$ is determined by beans in the training set and the Support Vector

Machine [15], one of the supervised machine learning algorithms. Roughly speaking, the training set is used to determine the model, while the test set is employed to see how effective the model is. As one can expect, the larger the training set is, the more accurate the model (1) is. In the existing literature, the ratio of the size of the training set to that of the test set is 7:3, and even up to 9:1 for obtaining higher accuracy of the model in some studies [17, 18]. On the other hand, the larger training set will lead to more cost computation. For our identifying scheme using six statistical characteristics, the accuracy rate is between 82% and 98% (see Table 3) for beans from different varieties or growing sites (see Section 2.1) as the ratio of the size of the training set to that of the test set is varied from 0.05 to 0.95 [19].

## 3. Results

### 3.1. A color feature of a qualified bean

To see the color feature of a qualified green coffee bean, we take 30 qualified green coffee beans from producer site **1,** as stated in **Materials and Methods**. For the red grayscale image corresponding to the red channel of the RGB image of a bean, we compute the red grayscale distribution curve showing the frequency of occurrence of each grayscale value found in this red grayscale image. Then the computed 30 red grayscale distribution curves, each corresponding to each of the 30 qualified beans, are depicted in Fig. 3(A). It can be seen that these red grayscale distribution curves have a high degree of similarity. Similarly, for these 30 qualified beans, their green (respectively, blue) grayscale distribution curves are computed and depicted in Fig. 3(B) and Fig. 3(C), respectively. Surprisingly, these green (respectively, blue) grayscale distribution curves are highly similar.

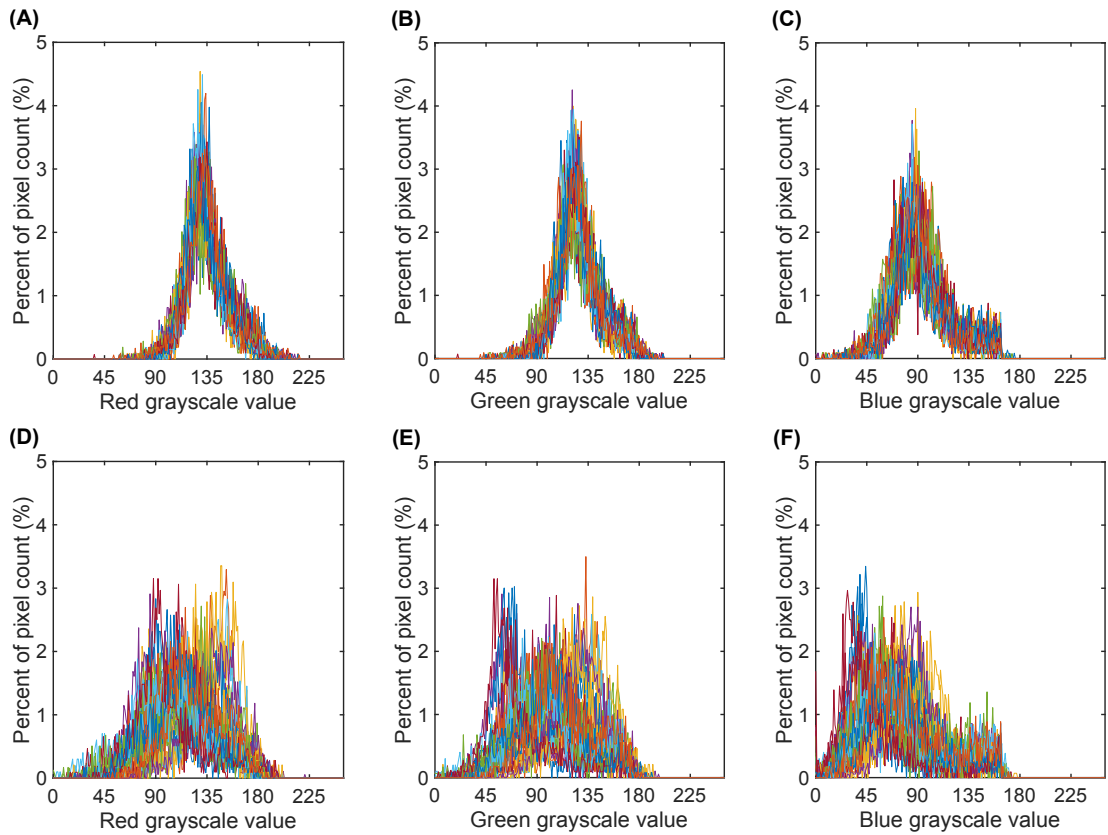

**Fig. 3 | Primary color grayscale distribution curves of 30 qualified/defective green coffee beans**: panels (A), (B), and (C) are respectively red, green, and blue grayscale distribution curves of qualified beans, while (D), (E), and (F) are respectively red, green and blue grayscale distribution curves of defective beans. Each curve in panels (A), (B), and (C) corresponds to a qualified bean, while each curve in panels (D), (E), and (F) corresponds to a defective bean. All of the beans are taken from producer site **1,** as stated in **Materials and Methods**.

On the other hand, we take 30 defective green coffee beans from producer site **1,** as stated in **Materials and Methods**, and then follow the above procedure to compute their respective primary color grayscale distribution curves. The obtained 30 red (respectively, green and blue) grayscale distribution curves are depicted in Fig. 3(D) (respectively, Fig. 3(E) and Fig. 3(F)). It can be observed from Fig. 3(D) that there are significant differences between the red grayscale distribution curves of different defective green coffee beans. Therefore, the degree of similarity of the red grayscale distribution curves of defective beans is significantly lower than that of qualified beans. Also, as shown in Fig. 3(E) and Fig. 3(F), similar conclusions hold for the green and blue grayscale distribution curves of defective beans. Furthermore, as shown in the **Supplementary materials**, beans from sites **1**, **2**, **3**, and **4** support the aforementioned conclusions.

These above results strongly indicate that any primary color grayscale value distribution curves of the seed coats of qualified green coffee beans from the same growing site should be ``nearly'' identical. This indicates that the primary color grayscale distribution curve of beans is a site-specific color feature of a normal bean. In the following sections, we will propose two evaluation schemes based on the statistics characteristics of qualified beans' primary color grayscale distribution curves. We note that the primary color grayscale value distribution of a normal bean cannot be fully determined by its basic statistics characteristics such as mean, variance and skewness. However, as we will see, the proposed evaluation schemes have very good accuracy rates, showing that the primary color grayscale distribution curves of qualified beans are truly site-specific color features.

### 3.2. Two-component statistical characteristics of qualified beans

To demonstrate the proposed color feature of qualified beans, we take 600 green coffee beans from producer site **1** stated in **Materials and Methods**, and divide them into two sets, as indicated below:

1. The training set consisted of 120 qualified beans and 120 defective beans, and
2. The test set consisted of 180 qualified beans and 180 defective beans.

Now, for each bean, compute the corresponding approximate normal distribution of its red grayscale distribution using maximum likelihood estimation. We note that the primary color grayscale distributions of the image of a qualified bean are not normal (Gaussian) distributions. We take its normal approximation to facilitate the analysis based on the machine learning classifier. Next, we find the mean ($x$) and standard deviation ($y$) of this approximate normal distribution. Note that the approximate normal distribution is fully determined by its mean and standard deviation. Now, the pair ($x, y$) is represented by a blue circle in the $xy$-plane if the bean from the test set is normal, while it is depicted as an orange ball if the bean from the test set is defective. In Fig. 4(A), it is observed that most of the blue circles fall below the separatrix line $L_r$, while most of the orange balls lie above $L_r$. Mathematically, the separatrix line $L_r$ can be expressed in the form

$$L_r: m_r x + s_r y - B_r = 0,$$

where the constants $m_r$, $s_r$ and $B_r$ can be computed using the SVM algorithm [15] (see **Materials and Methods**), and the subindex $r$ stands for red. Therefore, if the pair $(x, y)$ satisfies

$$C_{r1}: m_r x + s_r y - B_r < 0,$$

then the pair $(x, y)$ falls below the separatrix line $L_r$, and so the corresponding bean should be identified as a qualified bean. On the other hand, if the pair $(x, y)$ satisfies

$$C_{r2}: m_r x + s_r y - B_r > 0,$$

then the pair $(x, y)$ lies above the separatrix line $L_r$, and so the corresponding bean should be identified as a defective bean.

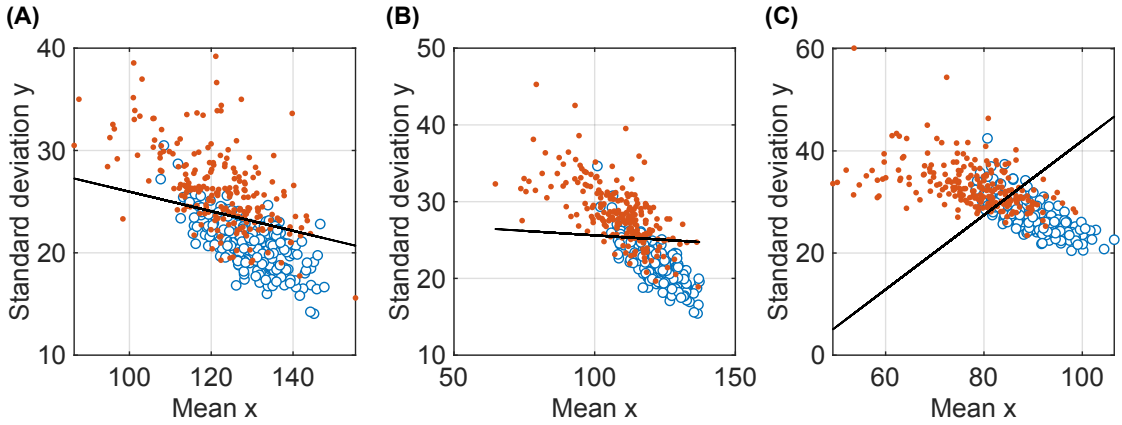

**Fig. 4 | Scatter plot of standard derivation versus mean for a primary color grayscale distribution of the image of beans**: (A) red grayscale distribution, (B) green grayscale distribution, and (C) blue grayscale distribution. An orange ball corresponds to a defective bean, while a blue circle corresponds to a qualified bean. The black solid line (denoted by $L_r$, $L_g$, and $L_b$ in panels (A), (B), and (C), respectively) almost separates orange balls from blue circles. The subindex $r$ (respectively, $g$ and $b$) means red (respectively, green and blue). Each point's horizontal (respectively, vertical) coordinate represents the mean (respectively, standard derivation) of the red grayscale distribution of the corresponding bean. The beans are taken from producer site **1** stated in **Materials and Methods**.

Motivated by the observations above, we calculate the following two quantities:
1. $PQ_r$ denotes the total number of qualified beans in the test set which satisfy the criterion $C_{r1}$;
2. $PD_r$ denotes the total number of defective beans in the test set which satisfy the criterion $C_{r2}$.

Here $PQ_r$ and $PD_r$ are the initials of predictable qualified and predictable defective, respectively. Thus, the sum of $PQ_r$ and $PD_r$ gives the total number of beans that the evaluation scheme can successfully classify. Then the *accuracy rate* of the scheme is defined by

$$\text{accuracy rate} = \frac{\text{PQ}_r + \text{PD}_r}{\text{total number of beans in the test set}}.$$

Therefore, the quantity *accuracy rate* measures the effectiveness of the scheme of classifying beans. For the current case, the accuracy rate is 82.22%, which shows that, in 82.22% of all beans in the test set, the evaluation scheme made the right classification. This, in turn, suggests that the proposed red color feature seems to be a good feature of qualified beans.

Before proceeding any further, we would like to make several remarks. First, the pair $(m_r, s_r)$ may depend on the producer sites of beans and the size of the training set of beans. However, as we will see later, the weight pair $(m_r, s_r)$ seems to be insensitive to the size of the training set. Second, instead of using the red grayscale distribution, we can apply the aforementioned evaluation scheme based on the green (respectively blue) grayscale distribution of beans to the beans in the test set, and the corresponding accuracy rate is 85.56% (respectively 83.89%). Therefore, any primary color statistics feature of beans also works well for classifying beans. Third, to show that the proposed primary color feature of beans is site-specific, we apply the evaluation scheme to beans from the growing sites **2-4** stated in **Materials and Methods**. The result is that except for the case of the blue grayscale distribution, the corresponding accuracy rate is at least 78.99% and as high as 96.95% for some cases (see Table 1 below). On the other hand, a more subtle identifying scheme, based on all of the three primary color grayscale distributions of beans and proposed in the coming section, can give an accuracy rate of at least 84.72% for beans from any growing sites. Nevertheless, the proposed statistical characteristics of any single primary color grayscale distribution is a good and simple characteristics for identifying qualified beans.

**Table 1 | The accuracy rate of the identifying scheme of green coffee beans**

| Growing site<br>Primary color | Site **1** | Site **2** | Site **3** | Site **4** |
|---|---|---|---|---|
| R | 82.22% | 81.16% | 84.44% | 96.95% |
| G | 85.56% | 78.99% | 80.00% | 93.13% |

| | B | 83.89% | 74.48% | 68.33% | 79.77% |

The result is based on single primary color grayscale distribution, for beans from different growing sites described in **Materials and Methods**. The ratio of the size of the training set to that of the test set is 4:6.

### 3.3. Six-component statistical characteristics of qualified beans

We will give an evaluation scheme with a higher accuracy rate based on the three primary color grayscale distributions. Indeed, for each bean, calculate the mean ($x_r$) (respectively, $x_g$ and $x_b$) and standard deviation ($y_r$) (respectively, $y_g$ and $y_b$) of the approximate normal distribution of its corresponding red (respectively, green and blue) grayscale distribution. Then a bean is assigned an ordered six-tuple $X = (x_r, x_g, x_b, y_r, y_g, y_b)$ in the six-dimensional space $\mathbb{R}^6$. As before, the six-tuples associated with qualified beans and those with defective beans are mostly separated by a hyperplane E which takes the form of

$$F(X) = \widehat{m_r}\, x_r + \widehat{m_g}\, x_g + \widehat{m_b}\, x_b + \widehat{s_r}\, y_r + \widehat{s_g}\, y_g + \widehat{s_b}\, y_b - B = 0,$$

where the constants $\widehat{m_i}$, $\widehat{s_i}$, $i = r, g, b$, and $B$ can be computed using the SVM algorithm. As before, we calculate the following two quantities:

1. PQ denotes the total number of qualified beans in the test set which satisfy the criterion $F(X) < 0$;
2. PD denotes the total number of defective beans in the test set which satisfy the criterion $F(X) > 0$.

Again, the sum of PQ and PD gives the total number of beans for which the scheme can successfully classify. Then the *accuracy rate* of the scheme is defined by

$$\text{accuracy rate} = \frac{PQ + PD}{\text{total number of beans in the test set}}$$

The accuracy rates of the scheme for different growing sites are given in Table 2 below. It can be seen that the accuracy rate for identifying beans is at least 84.72% and as high as 98.09%. Moreover, compared with the result from Table 1 for beans from the same growing site, the accuracy rate of the identifying scheme using three primary color grayscale distributions is higher than that using a single primary color one.

**Table 2 | The accuracy rate of the identifying scheme of green coffee beans for beans from different growing sites.**

| Growing site | Site 1 | Site 2 | Site 3 | Site 4 |
| --- | --- | --- | --- | --- |

| Accuracy | | | | |
|---|---|---|---|---|
| | 92.22% | 84.72% | 87.78% | 98.09% |

The ratio of the size of the training set to that of the test set is 4:6.

### 3.4. Site-specific property of the proposed characteristics of a bean

The two proposed evaluation schemes have good accuracy rates. However, when taking a look back at these schemes, one may raise the question: does the accuracy rate get higher as the ratio of the size of the training set to that of the test set is increased? At first glance, the answer is yes. This is because the larger the training set is, the more information one would get, so a more accurate separatrix hyperplane/curve for identifying beans can be determined. On the other hand, if the accuracy rate is shown to be high and not sensitive with respect to the ratio of the size between the training set and the test set, then this would suggest that our proposed color characteristic of a bean would be site-specific since less information will not reduce the accuracy rate and more information will not increase the accuracy rate.

To confirm this conjecture, we take 600 green coffee beans from producer site **1**, and investigate the dependence of the accuracy rate of the scheme $A$, using six characteristics, on the ratio of the size of the training set to that of the whole set of beans when this ratio is increased from 0.05 to 0.95. For comparison, we also compute the accuracy rate of scheme $S$ which uses each grayscale value in each pixel of the RGB image of a bean as the identifying characteristics [7]. Thus, the total number of the employed characteristics in scheme $S$ is $3 \times 256 = 768$ since each pixel has three grayscale values, and each grayscale value ranges from 0 to 255. It is observed from Fig. 5 that for the scheme $A$, the accuracy rate is almost the same and is above 90% as the ratio is increased. Therefore, the fact that the accuracy rate of our scheme $A$ is independent of the ratio size seems to indicate that our proposed color features of beans are indeed site-specific. On the other hand, for scheme $S$, the accuracy rate increases from 77% to 88% as the ratio is increased from 0.05 to 0.4, decreases as the ratio is increased from 0.4 to 0.5, and then increases as the ratio is increased from 0.5 to 0.8. Thus, the accuracy rate of scheme $S$ would depend on how much information we have in the training stage.

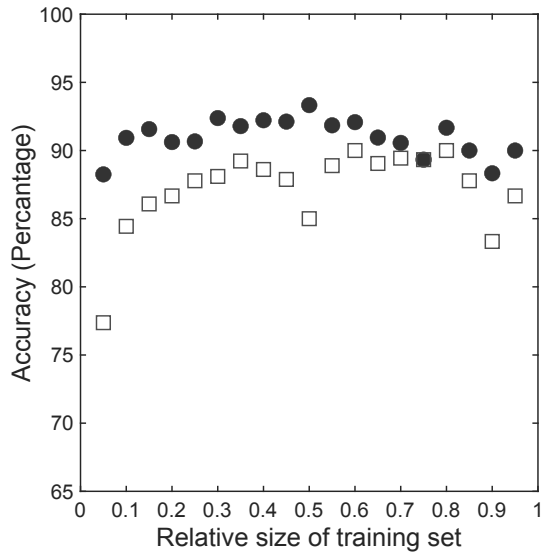

**Fig. 5 | The dependence of the accuracy rate on the ratio of the size of the training set to that of the test set.** The black ball corresponds to our proposed scheme using six color characteristics, while the square corresponds to the scheme which uses each grayscale value in each pixel of the RGB image of a bean as the identifying characteristic.

For beans from the other sites, we follow the procedure above to study the dependence of the accuracy rate of our proposed scheme $A$ on the ratio of the size of the training set to that of the whole set of beans. The result is given in Table 3 below, indicating that the accuracy rate of the evaluation scheme is quite robust to the ratio size. In particular, even the case with a small training set can give a high accuracy rate. So, our proposed color features of the image of beans are indeed site-specific.

**Table 3 | The accuracy of the identifying scheme for beans from different growing sites.**

| Growing site / Ratio | Site 1 | Site 2 | Site 3 | Site 4 |
|---|---|---|---|---|
| 0.05 | 88.25% | 82.63% | 82.28% | 97.10% |
| 0.10 | 90.93% | 84.26% | 88.70% | 98.21% |
| 0.15 | 91.57% | 85.29% | 87.65% | 98.38% |
| 0.20 | 90.62% | 86.46% | 86.46% | 97.99% |
| 0.25 | 90.67% | 85.78% | 87.11% | 97.24% |
| 0.30 | 92.38% | 85.00% | 87.14% | 98.04% |
| 0.35 | 91.79% | 85.64% | 86.92% | 97.89% |
| 0.40 | 92.22% | 84.72% | 87.78% | 98.09% |
| 0.45 | 92.12% | 83.33% | 88.79% | 97.92% |
| 0.50 | 93.33% | 85.67% | 88.00% | 99.54% |
| 0.55 | 91.85% | 83.70% | 88.89% | 97.96% |
| 0.60 | 92.08% | 82.92% | 87.92% | 98.28% |
| 0.65 | 90.95% | 84.29% | 87.14% | 98.68% |
| 0.70 | 90.56% | 83.33% | 86.67% | 98.46% |
| 0.75 | 89.33% | 84.00% | 86.00% | 97.22% |
| 0.80 | 91.67% | 85.83% | 85.83% | 98.86% |
| 0.85 | 90.00% | 87.78% | 85.56% | 98.48% |
| 0.90 | 88.33% | 83.33% | 86.67% | 97.73% |
| 0.95 | 90.00% | 93.33% | 83.33% | 100.00% |

The ratio of the size of the training set to that of the whole set of beans varies from 0.05 to 0.95.

## 3.5. Distinguish qualified beans from different growing sites

The site-specific property of qualified beans enables one to distinguish qualified beans from different growing sites. To see this, we apply the SVM machine learning classifier based on the site-specific features of qualified beans to a group of beans with their growing sites from sites **1**, **2**, and **3**. The confusion matrix of classification results is given in Fig. 6. Adding the three numbers in the top row of the confusion matrix gives the total sample beans from site **1** in the test set, 114+30+10=154. The leftmost panel of the top row indicates that 114 sample beans from site **1** are predicted to be from site **1**, so the model predicts correctly. The middle and rightmost panels of the top row indicate that 30 and 10 sample beans from site **1** are predicted to be from site **2** and site **3**, respectively, so the model's prediction is wrong. Similar explanations are for the other two rows. So, the correct classifications are the diagonal elements of the confusion matrix, and the accuracy rate for sample beans in the test set correctly classified is 87.29%. So, our proposed features of qualified beans can effectively distinguish qualified beans from different growing sites. This is an important application in the coffee industry, as some coffee distributors mix qualified beans from different production sites to lower costs.

**Fig. 6** |The confusion matrix of the trained model for distinguishing qualified beans from different growing sites.

## 4. Discussion and Conclusions

In this paper, we demonstrate that any specific primary color grayscale distribution curves of the seed coats of qualified beans from the same growing site, each of which shows the frequency of occurrence of each grayscale value found in this primary color grayscale image of the bean, have a very high degree of similarity and unimodal (see Fig. 7 and panels (A), (B) and (C) of Fig. 3). On the other hand, the primary color grayscale distribution curves associated with different defective green coffee beans are significantly different and polymodal (see Fig. 7 and panels (D), (E) and (F) of Fig. 3). These color characteristics of beans were verified to hold for beans from four different growing sites, including three types of Taiwan Arabica coffee beans and one type of India Robusta bean (see **Supplementary materials**). These evidences indicate that the primary color grayscale distribution curve is a site-specific color feature of a normal bean. This site-specific property of the color feature of qualified beans was also demonstrated by the machine learning classifier, the Support Vector Machine [15] (see Sec. 3.4). The summary of the main results is schematically shown in Fig. 7.

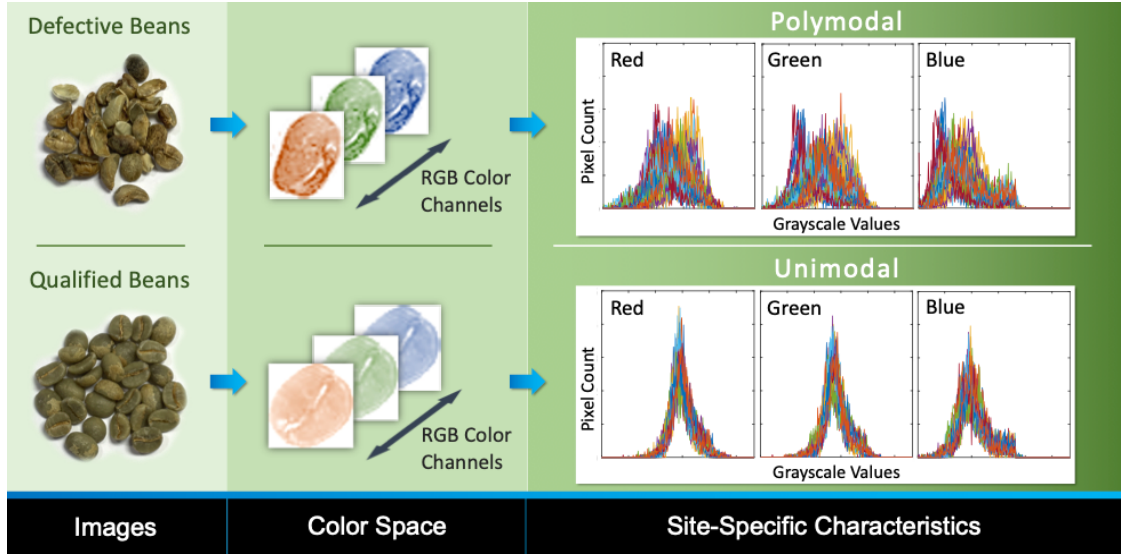

**Fig. 7 | The relation between qualified (respectively, defective) green coffee beans and their site-specific color features.**

Recall that in Sec. 3, we have introduced two evaluation/identifying schemes of beans: scheme *A*, based on the six statistics characteristics of all three primary color grayscale distribution curves of a bean, and scheme *B*, based on the two statistics characteristics of any single primary color grayscale distribution curve of a bean. Scheme *A* has a higher accuracy rate than scheme *B*. Moreover, the accuracy rates of these two schemes are insensitive to the ratio of the size of the training set to that of the test set, which indicates that the proposed features of a qualified bean are site-specific.

Compared with the existing methods, which require many training samples and the number of the chosen characteristics ranges from tens to tens of thousands, our scheme has the advantage of low-cost computation. For instance, deep learning methods such as convolutional neural networks have a high classification accuracy but require a huge amount of data for training [16]. Also, it is difficult to prepare sufficient amounts of sample beans as training data for a specific type of defective bean, particularly if the defect type is rare [16]. On the other hand, our evaluation/identifying schemes only need a small portion of sample beans for training. Indeed, as demonstrated in Tale 3, for a set of 1000 beans, a training set of 20 beans is all that is needed to achieve a classification accuracy higher than 82%. Another advantage of our evaluation/identifying scheme is site-specific, meaning that two different sets of beans from the same growing sites and varieties share the same color characteristics. However, for two different groups of beans from the same growing sites and types, the extracted color characteristics by the typical machine learning classifiers may be different [14].

## Acknowledgments

This work is supported by NSTC with the grant number 111-2813-C-007-105-M.

## Conflict of Interest

The authors declare no conflict of interest.

## Declaration

The green coffee beans used in this study were granted permission from local farmers or coffee distributors in Taiwan for academic research.

## Data availability

The data that support the findings of this study are available on request from the corresponding author.

## Supporting Information

Supplementary materials are available, which contain (1) the results for beans from sites **2**, **3**, and **4**; and (2) a short description of the common defective beans.